# News Headlines Dataset For Sarcasm Detection

**Rishabh Misra**
UC San Diego
r1misra@eng.ucsd.edu

**Abstract.** Past studies in Sarcasm Detection mostly make use of Twitter datasets collected using hashtag-based supervision but such datasets are noisy in terms of labels and language. Furthermore, many tweets are replies to other tweets, and detecting sarcasm in these requires the availability of contextual tweets. To overcome the limitations related to noise in Twitter datasets, we curate News Headlines Dataset from two news websites: TheOnion aims at producing sarcastic versions of current events, whereas HuffPost publishes real news. The dataset contains about 28K headlines out of which 13K are sarcastic. To make it more useful, we have included the source links of the news articles so that more data can be extracted as needed. In this paper, we describe various details about the dataset and potential use cases apart from Sarcasm Detection.

## 1. Limitations of Existing Datasets

There have been many works in Sarcasm Detection in the past that have used either a small high-quality labeled dataset or a large noisily labeled dataset. We will cover some of the prominent works in this section to motivate the need for our dataset. (Amir et al., 2016) use a large-scale Twitter-based dataset collected using hashtag-based supervision. They propose to use a CNN to automatically extract relevant features from tweets and augment them with user embeddings to provide more contextual features during sarcasm detection. However, the dataset is limited in the following aspects:
- The Twitter dataset used in the study was collected using hashtag-based supervision. As per various studies [(Liebrecht et al., 2013; Joshi et al., 2017)], such datasets have noisy labels.
- Furthermore, people use very informal language on Twitter which introduces sparsity in vocabulary, and for many words pre-trained embeddings are not available.

- Lastly, many tweets are replies to other tweets, and detecting sarcasm in these requires the availability of contextual tweets.

On the other hand, Semeval Challenge[1] released a Twitter-Based dataset where the tweets are manually labeled. This helps with removing label noise but at least the second problem reported above may still exist. Additionally, due to manual labeling, the scale of the dataset is very small so it is unlikely that any modern Deep Learning approach will work well.

## 2. News Headlines Dataset

To overcome the limitations related to noise in Twitter datasets, we collected a New Headlines Dataset[2] from two news websites. TheOnion[3] aims at producing sarcastic versions of current events and we collected all the headlines from "News in Brief" and "News in Photos" categories (which are sarcastic). We collected real (and non-sarcastic) news headlines from HuffPost[4]. The general statistics of this dataset along with the dataset provided by the Semeval challenge are given in Table 1. We can notice that for the Headlines dataset, where text is much more formal in language, the percentage of words not available in word2vec vocabulary is much less than Semeval dataset.

| Statistic/Dataset | Headlines | Semeval |
| --- | --- | --- |
| # Records | 28,619 | 3,000 |
| # Sarcastic records | 13,635 | 2,396 |
| # Non-sarcastic records | 14,984 | 604 |
| % of pre-trained word embeddings not available | 23.35 | 35.53 |

Table 1: General statistics of the datasets.

This new dataset has the following advantages over the existing Twitter datasets:
- Since news headlines are written by professionals in a formal manner, there are no spelling mistakes and informal usage. This reduces the sparsity and also increases the chance of finding pre-trained embeddings.
- Furthermore, since the sole purpose of TheOnion is to publish sarcastic news, we get high-quality labels with much less noise as compared to Twitter datasets.
- Unlike tweets which are replies to other tweets, the news headlines we obtained are self-contained. This would help methods to tease apart the real sarcastic elements.

---

[1] https://competitions.codalab.org/competitions/17468
[2] Dataset is available at https://rishabhmisra.github.io/publications/
[3] https://www.theonion.com/
[4] https://www.huffpost.com/

Each record in the dataset consists of three attributes:
- `is_sarcastic`: 1 if the record is sarcastic otherwise 0
- `headline`: the headline of the news article
- `article_link`: link to the original news article.

We include an article link corresponding to each headline so that more text regarding the news can be extracted as needed by any Machine Learning task.

## 3. Data Curation Method

We make use of open-source tools like BeautifulSoup, Selenium, and Chrome Driver to curate the dataset. For collecting data from TheOnion, we use News in Brief and News in Photos as the base link. In all the headlines presented, we extract their link and headline text using BeautifulSoup API. Once that is done on one page, we simulate a button click action using Selenium to go to the next page and repeat the process. We combine the data from these two categories into one. This gives us the Sarcastic version of the dataset. For the non-sarcastic part, we treat Huffington Post's archive link as the base and extract headlines following the same procedure. Since Huffington has significantly more news headlines, we downsample the data to approximately match the number of sarcastic headlines. We then do vertical integration of data and shuffle all the records. Since the headline text comes from professional news websites and has reasonably good quality (no misspellings, abbreviations, etc.), we did not do any additional pre-processing on it.

## 4. Reading the Data

Once you download the dataset, you can use the following code snippet to read the data for your machine learning methods:

```python
import json

def parse_data(file):
    for l in open(file,'r'):
        yield json.loads(l)

data = list(parse_data('./Sarcasm_Headlines_Dataset.json'))
```

## 5. Exploratory Data Analysis

As a basic exploration, we visualize the word clouds for the sarcastic and non-sarcastic categories in Figures 1 & 2 respectively, through which we can see the types of words that occur frequently in each category.

Figure 1: Word Cloud from Sarcastic News Headlines.

Figure 2: Word Cloud from Non-Sarcastic News Headlines.

On the surface level, we don't see any difference in the types of words used within each category. We furthermore look at text length and the sentiment contained within the headline in Figure 3. There does not appear to be any describable difference between the two classes.

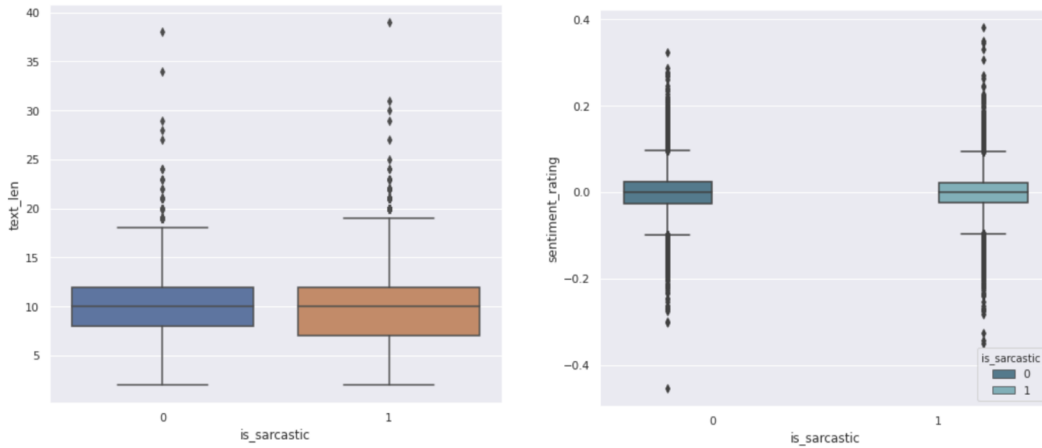

Figure 3: Text length and sentiment for sarcastic (1) and non-sarcastic text (0)

All of this may confirm that Sarcasm Detection may involve more subtle differences in the text and may need special modeling work.

## 6. Potential Use Cases

The "News Headlines Dataset" can also be thought of as a collection of real and fake news and would serve as a great source to train machine learning models to track down fake news on the Internet. With the proliferation of fake news around any major global events (like elections and pandemics), the tool trained on this data can be invaluable Furthermore, the dataset can also be used to study humor and irony-related linguistic phenomena which are again difficult to model. As a starting point for explorations, Misra et. al. provide a code implementation[5] of using the dataset for tackling a learning task in the PyTorch framework.

## References

[1] Amir, Silvio, Byron C. Wallace, Hao Lyu, and Paula Carvalho Mário J. Silva. "Modelling context with user embeddings for sarcasm detection in social media." *arXiv preprint arXiv:1607.00976*.

---

[5] https://github.com/rishabhmisra/Sarcasm-Detection-using-NN